# Machine Unlearning for Causal Inference

Vikas Ramachandra[1] & Mohit Sethi[2]

Associate Professor and CTO[1]

[vikas.ramachandra@jacobs.ucsd.edu](vikas.ramachandra@jacobs.ucsd.edu)[1]

[m.sethi006@gmail.com](m.sethi006@gmail.com)[2]

**Abstract:**

Machine learning models play a vital role in making predictions and deriving insights from data. However, in some cases, the deployment of these models can have unintended consequences or introduce biases. In the field of causal analysis [1], where understanding cause-and-effect relationships is crucial, it becomes essential to have models that accurately capture causal relationships. At the same time, to preserve user privacy, it is important to enable the model to 'forget' some of its learning/captured information about a given user. This is where machine unlearning is useful. Additionally, machine unlearning can aid in handling concept drift, where models adapt to changing data distributions over time, by allowing them to unlearn outdated patterns and learn from more recent data.

This paper introduces the concept of machine unlearning [4] for causal inference, particularly propensity score matching and treatment effect estimation, which aims to refine and improve the performance of machine learning models for causal analysis [1] given the above unlearning requirements.

In the field of causal analysis, understanding cause-and-effect relationships is vital, and accurately capturing these relationships requires models that can handle biases and unwanted associations effectively. Additionally, ensuring user privacy is equally important, and enabling models to "forget" certain user-specific information is essential to comply with privacy regulations. Therefore, the motivation behind this research is to propose a novel machine-unlearning methodology that preserves user privacy.

The paper presents a methodology for machine unlearning [4] using a neural network-based propensity score model. The dataset used in the study is the Lalonde dataset [7], a widely used dataset for evaluating the effectiveness i.e. the treatment effect of job training programs. The methodology involves training an initial propensity score model on the original dataset and then creating forget sets by selectively removing instances, as well as matched instance pairs. based on propensity scores. These forget sets are used to evaluate the retrained model, allowing for the elimination of unwanted associations. The actual retraining of the model is performed using the retain set.

The experimental results demonstrate the effectiveness of the machine unlearning [4] approach. The distribution and histogram analysis of propensity scores before and after unlearning provide insights into the impact of the unlearning process on the data.

This study represents the first attempt to apply unlearning techniques to causal inference. The findings of this study highlight the potential of machine unlearning [4] techniques for refining and enhancing the accuracy of machine unlearning [4] in causal analysis [1]. The proposed methodology offers a framework for addressing biases and improving causal inference [2] in various domains. Further research and exploration of machine unlearning [4] techniques hold promising avenues for improving the reliability and fairness of machine learning models in causal analysis [1].

## Introduction

Machine learning algorithms have become a cornerstone in data-driven decision-making across various domains. These algorithms are designed to identify patterns, make predictions, and extract insights from large datasets. However, their deployment can sometimes lead to unintended consequences, such as biased predictions or the amplification of existing societal disparities. This is particularly critical in the field of causal analysis [1], where accurately identifying cause-and-effect relationships is essential for making informed decisions and implementing effective interventions.

The study of causal relationships often relies on propensity score models, which estimate the likelihood of treatment assignment given a set of covariates. These models help researchers account for confounding factors and mitigate biases when estimating causal effects. However, even with the use of propensity scores, machine learning models can still exhibit biases and associations that may impact the accuracy of causal inference [2].

To address these challenges, this paper introduces the concept of "machine unlearning" on causal datasets. Machine unlearning [4] is an iterative process of refining and improving machine-learning models specifically tailored for causal analysis [1]. The goal is to identify and eliminate biases and unwanted associations in the data, thereby enhancing the accuracy and reliability of causal inferences [2].

The Lalonde dataset [7], a well-known benchmark dataset evaluating the effectiveness of job training programs, serves as the basis for the experiments conducted in this study. The dataset consists of various covariates, such as age, education level, and income, along with a binary treatment indicator representing participation in the training program. By applying the machine unlearning [4] methodology to the Lalonde dataset [7], the study aims to demonstrate the efficacy of this approach in demonstrating machine unlearning for causal inference [2].

The proposed methodology involves training an initial propensity score model on the original dataset. Subsequently, forget sets are created by selectively removing instance pairs based on propensity score matching. The retain sets are then used to retrain the model, allowing for the identification and elimination of biases introduced by the original model. By comparing the performance of the retrained models with the original model using metrics such as root mean squared error (RMSE), the impact of machine unlearning [4] on causal analysis [1] can be assessed.

In addition to evaluating model performance, this paper also analyzes the distribution and histogram of propensity scores before and after the unlearning process. This analysis provides insights into the changes in the associations and biases present in the data and further supports the effectiveness of the machine unlearning [4] approach.

The findings of this study have important implications for researchers and practitioners in the field of causal analysis [1]. By leveraging machine unlearning [4] techniques, it becomes possible to refine machine-learning models while retaining causal information. This can lead to more reliable and fair decision-making processes, ultimately contributing to the advancement of evidence-based interventions and policies. The ability to "forget" certain information while retaining the essence of causal relationships is a groundbreaking capability offered by machine unlearning. As a result, privacy concerns can be effectively addressed without compromising the integrity of causal analyses. This feature is of particular importance when working with sensitive data, ensuring compliance with privacy regulations, and fostering trust among users and stakeholders.

Moreover, the potential applications of machine unlearning extend beyond causal analysis, offering diverse avenues for enhancing machine learning models' reliability in various domains. Concept drift, a common challenge in dynamic environments, can be tackled with ease by allowing models to adapt to new patterns without being constrained by outdated data, thus ensuring up-to-date and accurate predictions over time.

In sum, the methodology introduced in this study marks the first attempt at applying unlearning to causal inference [1]. By embracing these novel techniques, researchers and practitioners can evaluate the quality of causal analyses, thus empowering evidence-driven decision-making, and positively impact society by promoting fairness, transparency, and trust in machine learning models. As the adoption of machine unlearning continues to evolve, we anticipate its profound contributions to shaping a more equitable and informed world.

Overall, this paper contributes to the growing body of research on machine unlearning [4] and its application in the context of causal analysis [1]. By showcasing the potential of this approach, it encourages further exploration and development of machine unlearning [4] techniques to address biases and improve causal inference [2] in various domains.

## Datasets & Methods:

This paper utilizes the Lalonde dataset [7], a widely-used benchmark for evaluating job training program effectiveness, to investigate the application of machine unlearning [4] in causal analysis [1]. The dataset consists of various covariates, such as age, education level, race, and income, collected from a sample of individuals. The treatment variable indicates participation in the job training program, while the outcome variable is the post-program income.

To preprocess the data, standardization is applied to the numerical covariates using the StandardScaler from the scikit-learn library [6]. This step ensures that the covariates are on a similar scale, preventing any undue influence from variables with larger magnitudes.

The propensity scores, which estimate the probability of treatment assignment based on the covariates, are generated using a neural network-based model. This approach is borrowed from our earlier work, "Deep Learning for Causal Inference" [8]. The model architecture includes multiple linear layers with ReLU activation functions, culminating in a sigmoid activation function to produce the propensity scores. Training is performed using binary cross-entropy loss and the Adam optimizer.

The initial propensity score model is trained on the original dataset, using the covariates as inputs and the treatment indicator as the target variable. This model serves as the benchmark for comparison.

To create the forget set, two approaches are employed. The first approach, propensity matching pair-wise removal, involves iteratively selecting pairs of instances—one from the treatment group and one from the control group—based on their propensity scores. The nearest neighbor algorithm identifies the closest control instance for each treatment instance, and the selected pairs are added to the forget set. The instances in the forget set are removed from consideration for retraining the model. The second approach involves randomly selecting a specified percentage of instances from both the treatment and control groups to form the forget set, without propensity score matching.

The model is then retrained using the instances in the retain set, which comprises the instances not included in the forget set. The retraining process helps identify and eliminate biases and unwanted associations present in the original model. The same neural network architecture and training procedure as the initial model are utilized.

The retrained models, denoted as Model 2 and Model 3, are evaluated using the root mean squared error (RMSE) metric to assess their performance in estimating the treatment effect. The original model, trained on the entire dataset, serves as the benchmark for comparison.

To gain insights into the changes introduced by the machine unlearning [4] process, the distribution and histogram of propensity scores are analyzed before and after retraining the models. Kernel density estimation and histogram plots provide visual representations of the propensity score distributions, aiding in the identification of shifts in associations and biases in the data.

By combining forget set creation, model retraining, and propensity score analysis, this study explores the effectiveness of machine unlearning [4] in refining propensity score models for improved causal inference [2].

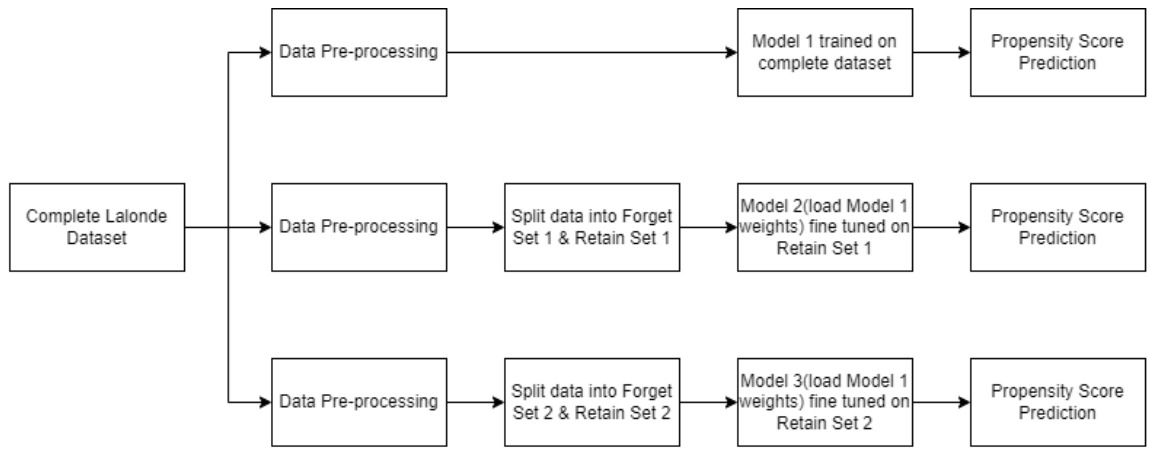

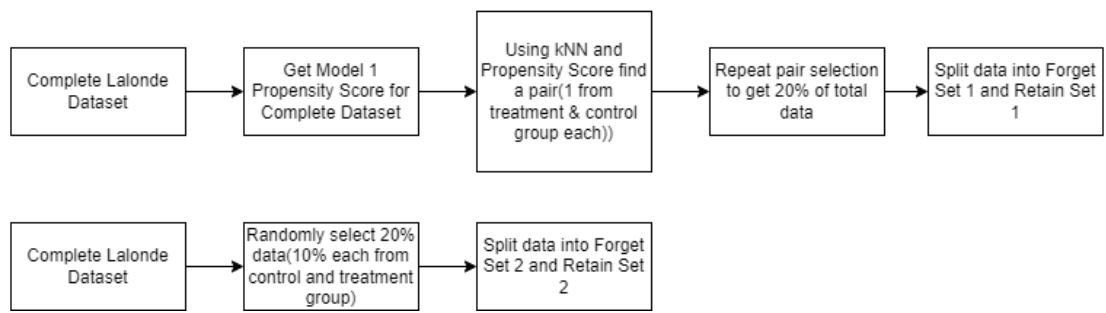

## Results:

When analyzing the propensity score distributions, it is observed that forget set 1, created using propensity matching pair-wise removal, demonstrates less overlap between the distributions compared to the original dataset. This finding suggests that the pair-wise removal approach in machine unlearning [4] effectively reduces biases and associations present in the original model.

In contrast, forget set 2, generated by random removal of instances, exhibits almost complete overlap in the distribution of propensity scores compared to the original dataset. This indicates that the random removal approach does not significantly alter the associations and biases in the model.

The evaluation of the propensity score models on the original dataset reveals the following root mean squared error (RMSE) values: Model 1 - 0.0579, Model 2 - 0.1087, and Model 3 - 0.0592. Model 1, trained on the entire dataset, performs the best with the lowest RMSE. Model 2, retrained using propensity matching pair-wise removal on the retrain set, shows a higher RMSE compared to the original dataset. Model 3, retrained on the retrain set created by random removal of instances, exhibits a similar RMSE to Model 1.

The similarity in RMSE values between Model 1 and Model 3 on the original dataset suggests that the random removal approach does not result in substantial unlearning. On the other hand, the lower RMSE value for Model 1 compared to Model 2 indicates that the propensity matching pair-wise removal approach leads to a better unlearning outcome.

Based on these results, it can be concluded that the pair-wise removal approach using propensity score matching [5] is more effective in machine unlearning [4] for causal analysis [1]. This approach successfully reduces biases and associations present in the original model, as evidenced by the reduced overlap in the propensity score distributions. Overall, these findings highlight the potential of machine unlearning [4] techniques in refining propensity score models and improving causal inference [2].

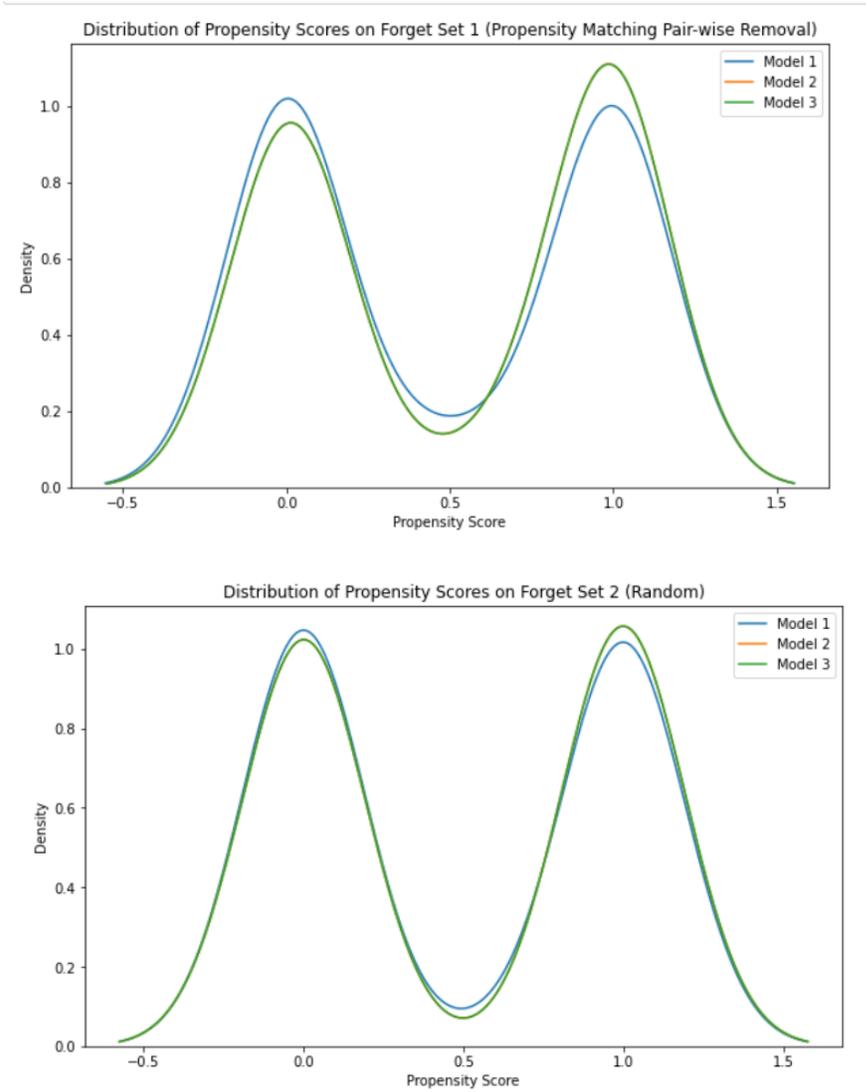

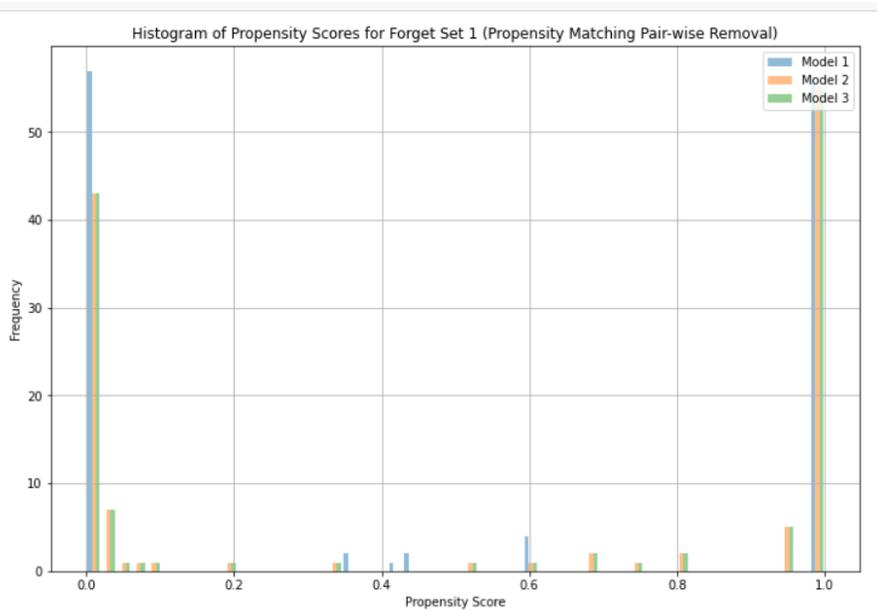

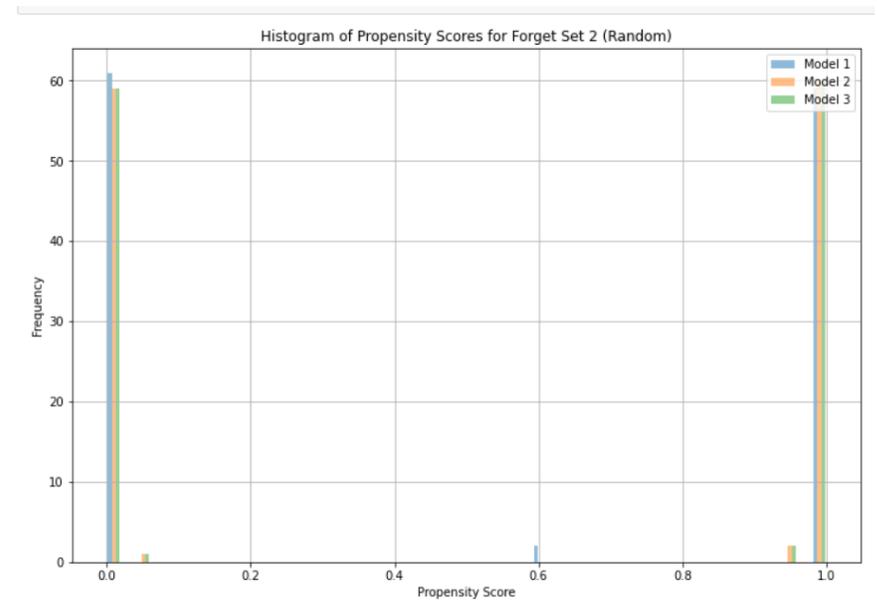

## Conclusion:

In this study, we explored the application of machine unlearning [4] techniques on a causal dataset using propensity score models. The evaluation of different unlearning approaches and their impact on the model's performance provided valuable insights into the effectiveness of these techniques.

Our findings indicate that pair-wise removal using propensity score matching [5] offers a superior approach to machine unlearning [4] for causal inference/treatment effect estimation. This approach resulted in a better unlearning outcome, as indicated by the lower RMSE values on the original dataset. Also, the lesser overlap between propensity score distribution on the forget set 1 compared to the original dataset indicates that the model has successfully unlearned.

In contrast, the random removal approach showed minimal changes in the associations and biases present in the model, as indicated by the almost complete overlap in propensity score distributions in forget set 2 compared to the original dataset. The RMSE values between the original dataset and the model retrained on forget set 2 were comparable, suggesting limited unlearning.

These findings emphasize the importance of carefully selecting the unlearning approach in machine unlearning [4] tasks. Propensity matching pair-wise removal offers a promising strategy for effectively reducing biases and associations in the model, leading to improved causal inference [2].

Overall, our study contributes to the growing body of research on machine unlearning [4] and its applications in causal analysis [1]. It highlights the potential of machine unlearning [4] techniques in refining models, enhancing causal inference [2], and promoting the robustness and reliability of causal studies. Further research can focus on exploring other unlearning techniques, investigating their effectiveness on different types of datasets, and examining their impact on various causal inference [2] tasks.